\documentclass[onecolumn]{svjour3}

\usepackage[margin=1in]{geometry} 
\usepackage{color} 
\usepackage{amsmath} 
\usepackage{amsfonts} 
\usepackage{graphicx} 
\usepackage{bm} 
\usepackage[utf8]{inputenc}
\usepackage{authblk} 
\usepackage{multicol} 
\usepackage{booktabs} 
\usepackage{lineno} 
\usepackage{caption} 
\usepackage[numbers,sort&compress]{natbib}

\journalname{Journal of the Brazilian Society of Mechanical Sciences and Engineering}

\begin{document}


\title{Artificial Intelligence in Mechanical Engineering: Foundations, Opportunities, Challenges, and Responsibilities}

\titlerunning{}














\author{
Americo~Cunha~Jr*\thanks{*Corresponding author.} \and 
C. Nataraj \and 
Luciano M. Lorenzo \and
\linebreak Hervé Panetto \and 
Americo Scotti \and 
José V. C. Vargas
}

\authorrunning{
A. Cunha~Jr \and 
L. M. Lorenzo \and 
H. Panetto \and 
A. Scotti \and 
J. V. C. Vargas}

\institute{A. Cunha~Jr \at
            ORCID: 0000-0002-8342-0363\\
              National Laboratory of Scientific Computing, Av. Getúlio Vargas, 333, Petrópolis, 25651-075, RJ, Brazil \\
              Rio de Janeiro State University,
			  Rua S\~{a}o Francisco Xavier, 524, Rio de Janeiro, 20550-900, RJ, Brazil\\
              \email{americo@lncc.br}
            \and
            C. Nataraj \at 
            ORCID: 0000-0002-5816-3932\\
            Moritz Endowed Professor, Villanova Center for Analytics of Dynamic Systems (VCADS), Villanova University, Villanova, PA 19085, USA \\
            \email{nataraj@villanova.edu}
            \and
            L. M. Lorenzo \at 
            Dassault Systèmes, Av. Jornalista Roberto Marinho, 85, São Paulo, 04576-010, SP, Brazil\\
            \email{luciano\_lorenzo@yahoo.com}
            \and
            H. Panetto \at 
            ORCID: 0000-0002-5537-2261\\
            University of Lorraine, CNRS, CRAN, Nancy, F-54000, France\\
            \email{herve.panetto@univ-lorraine.fr}
            \and
            A. Scotti \at 
            ORCID: 0000-0002-1005-5895\\
            Federal University of Uberlândia, Rodovia BR-050, km 78 - Bloco 1DCG, Uberlândia, 38410-337, MG, Brazil\\
            \email{ascotti@ufu.br}
            \and
            J. V. C. Vargas \at
            ORCID: 0000-0002-1458-2908\\
            Federal University of Paraná, CP 19011, Curitiba, 81531–980, PR, Brazil\\
            \email{viriato@ufpr.br}
}

\date{Received: date / Accepted: date}

\maketitle

\begin{abstract}

Artificial intelligence (AI) is rapidly transforming mechanical engineering (ME) practice, influencing activities that range from modeling and simulation to optimization, diagnostics, design, and technical communication. This article consolidates perspectives discussed at a dedicated round-table during the 28th ABCM International Congress on Mechanical Engineering — COBEM 2025 — and offers a structured reflection for the Brazilian and international ME communities. Rather than providing an exhaustive survey, the paper develops a coherent narrative around the conditions under which AI can strengthen---rather than dilute---engineering rigor. The article begins by clarifying how contemporary AI systems operate, distinguishing statistical learning from human cognition and outlining their capabilities, limitations, and emerging agentic AI systems. It then examines industrial adoption, workforce transformation, and the growing role of AI in scientific discovery and engineering workflows. Particular attention is devoted to the energy and infrastructure demands of AI-driven computation and their implications for sustainability, power systems, and thermal management. Building on these constraints, the paper discusses integrating physical knowledge, structured reasoning, and domain expertise into AI systems for safety-relevant engineering applications. Emphasis is placed on representative ME applications, physics-informed and hybrid learning frameworks, model development under physical constraints, validation, uncertainty quantification, and certification challenges. The paper also addresses the educational, epistemological, and ethical implications of AI, including curriculum modernization, human-agent collaboration, professional accountability, and the long-term sustainability of engineering knowledge. It concludes with a synthesis of principles for the responsible adoption of AI in ME, encompassing physical consistency, validation, awareness of uncertainty, energy sustainability, human oversight, and ethical responsibility. Our central message is that AI should be viewed not as a substitute for engineering judgment, but as a powerful tool whose value depends on disciplined integration with physical reasoning, scientific rigor, and professional responsibility.

\end{abstract}


\section{Introduction}
\label{sec1}

Artificial intelligence (AI) is rapidly reshaping the practice of mechanical engineering (ME), from early-stage concept generation and requirements analysis to modeling and simulation, optimization, manufacturing planning, control, maintenance, and technical communication. In particular, the mainstream availability of generative AI has lowered the barrier for engineers to interact with complex knowledge bases through natural language, and has accelerated adoption in both academic and industrial settings
\cite{Baker_LLM_ME_2025,Wang2023}. These capabilities are increasingly embedded within broader digital engineering frameworks (digital twins, digital shadows, and digital models) where data-driven models, physics-based simulations and emulations, and decision support coexist throughout the system life cycle \cite{Chakraborty2021p106410,Kapteyn2020DT,Silva2025DigitalTwin,Ganguli2020p1110,Ganguli2023,NAE2024DigitalTwins,Ritto20211p07614}. Recent global evidence also suggests that organizations are actively redesigning workflows and elevating governance structures to capture value from generative AI, reflecting both the significant promise of these technologies and the fact that many deployments remain at an early stage of maturity \cite{McKinsey_State_of_AI_2025}.

Nevertheless, AI is opening important opportunities across the full spectrum of ME practice. In research and advanced engineering, it can accelerate design-space exploration, support surrogate modeling for computationally intensive simulations, improve diagnostics from large sensor datasets, and enable new forms of interaction with models and technical knowledge. In day-to-day professional settings---including factories, maintenance routines, production engineering, quality assurance, and engineering management---AI can support routine but essential activities such as data analysis, report preparation, anomaly detection, process monitoring, and technical decision support. These possibilities help explain the growing interest in AI and clarify why its integration into ME deserves careful but constructive attention. This constructive perspective is reinforced by broader evidence that AI is beginning to generate measurable productivity gains across a variety of organizational settings, while also lowering barriers to advanced computational capabilities through more affordable and accessible models \cite{stanford_ai_index_2025}.

This momentum, however, arrives with a distinctive set of constraints and responsibilities in ME. Unlike domains in which outputs are primarily informational, ME produces artifacts and decisions embedded in the physical world, where safety, reliability, compliance, traceability, and accountability are non-negotiable. As highlighted in the digital twin literature, integrating AI into engineering systems therefore requires explicit attention to model validity, life-cycle consistency, and system-level interoperability \cite{Semeraro2021p103469}. The question is therefore not whether AI will be used, but under what conditions it can be used constructively—i.e., in ways that are verifiable, robust in the face of uncertainty, and aligned with professional responsibility.

A second, often underestimated constraint is energy sustainability. AI-enabled workflows scale through computation, and computation scales through energy and infrastructure. The growing electricity demand from data centers and AI workloads has become material enough to feature in dedicated energy-system analyses, sharpening the need to connect “responsible AI” not only to ethics and safety, but also to energy, carbon, and water footprints \cite{IEA_Energy_and_AI_2025}. For mechanical engineers—whose field has historically been central to energy conversion, thermal management, and efficiency—this creates an opportunity and an obligation: to help design AI-enabled systems that are not only powerful but also energy-aware and sustainable.

Even so, the institutional environment around AI is changing quickly. In 2023–2024, governance and risk-management guidance evolved from voluntary ``best practices'' into increasingly formalized frameworks and regulation, including the NIST AI Risk Management Framework (AI RMF 1.0), ISO/IEC guidance on AI risk management, and the EU Artificial Intelligence Act \cite{NIST_AIRMF_2023,ISO_IEC_23894_2023,EU_AI_Act_2024_1689}. While these instruments are not engineering standards \emph{per se}, they reinforce expectations that matter directly to engineering workflows: transparency about intended use, life-cycle risk management, monitoring and incident response, and appropriate human oversight—primarily when AI is used in high-consequence contexts.

In Brazil, the Brazilian Plan for AI (PBIA 2024--2028), in Portuguese \emph{Plano Brasileiro de Inteligência Artificial}, provides an important national framework for AI development, combining investments in computational infrastructure, research capacity, workforce formation, public services, and technological sovereignty. The plan highlights the strategic role of high-performance computing, scientific research, and education in enabling responsible and competitive AI development within the country \cite{PBIA2024}.

Similarly, the European Union has recently launched the AI Continent Action Plan  \cite{EUAIContinent2025}, which complements the regulatory framework established by the EU Artificial Intelligence Act \cite{EU_AI_Act_2024_1689}. Beyond regulation, the initiative emphasizes strategic investments in high-performance computing infrastructure, AI factories, data infrastructure, workforce development, and industrial innovation, with the objective of strengthening European technological sovereignty and accelerating the responsible deployment of AI across key sectors.

Recent developments have also expanded AI beyond predictive and generative models toward agentic systems capable of orchestrating complex engineering workflows that involve simulations, databases, optimization routines, and digital twins. These developments raise new questions concerning validation, uncertainty quantification, human oversight, and the transferability of AI models from simulation environments to real-world engineering systems.

Motivated by these developments, this document consolidates perspectives discussed on the Round Table on Artificial Intelligence that happened at the 28th ABCM International Congress on Mechanical Engineering --- COBEM 2025\footnote{This event was held in Curitiba, PR, Brazil, between November 9 --15, 2025.}, moderated by Prof. J. V. C. Vargas and Prof. A. Cunha Jr, with the participation of four scholars (I. Dincer, C. Nataraj, H. Panetto, and A. Scotti) and two professionals from industry (A. Nabholz and L. M. Lorenzo). Rather than attempting an exhaustive survey, our goal is to offer a trustworthy, practice-oriented synthesis for the Brazilian and international ME communities: \textit{how to adopt AI in ways that strengthen engineering quality and judgment, while mitigating risks and avoiding superficial ``automation optimism''.}

The contributions assembled here form a deliberate narrative. In Section~\ref{sec2}, A. Cunha~Jr. offers an opening overview of how contemporary AI systems operate, outlining the main learning paradigms, architectural foundations, and current capabilities, while clarifying limitations and distinguishing statistical learning from human cognition. The discussion also introduces emerging agentic AI systems, in which large language models are coupled with planning, tool use, and interaction with external software, highlighting their growing relevance to scientific discovery and engineering workflows. This establishes a common conceptual ground for the sections that follow, avoiding both technological hype and undue skepticism.

Section~\ref{sec3} brings an industry perspective by L. M. Lorenzo, who discusses how organizations are adopting AI and reorganizing governance, processes, and workforce skills around it, while noting that many current deployments remain concentrated in applications involving text, image, and code generation. In Section~\ref{sec4}, J. V. C. Vargas addresses a critical but often under-emphasized constraint on large-scale AI deployment: the energy infrastructure required to sustain AI-driven computation. His contribution examines the rapidly growing electricity demand of data centers, the associated challenges and opportunities related to power density, cooling, grid capacity, and sustainability, and the implications for ME. By discussing alternative energy pathways and system-level energy integration, the section highlights that responsible AI adoption is inseparable from energy efficiency, resource availability, and environmental impact.

This energy- and reliability-constrained context motivates Section~\ref{sec5}, a technical perspective by H. Panetto, which argues that, to be viable in ME, AI approaches must move beyond purely data-driven pattern recognition and explicitly integrate structured knowledge, reasoning, and physical consistency. The discussion points to hybrid methodologies aimed at improving generalization under sparse or noisy data, enhancing explainability, and building trust in safety-relevant applications. In Section~\ref{sec6}, C. Nataraj develops these ideas in the specific setting of model development and diagnostics for dynamic systems, emphasizing the ill-posed inverse nature of many engineering inference problems and the role of physics-informed, hybrid, and multi-fidelity learning frameworks in achieving robustness, interpretability, and long-horizon predictive fidelity, particularly in digital-twin settings.

To connect these conceptual discussions with current engineering practice, Section~\ref{sec7}, authored by A. Cunha~Jr., presents representative applications of AI in ME. The section surveys areas such as structural health monitoring, fatigue and fracture mechanics, additive manufacturing, computational mechanics, materials and metamaterials design, predictive maintenance, and emerging agentic engineering workflows. Particular emphasis is placed on the integration of data-driven methods with physical modeling, simulation, uncertainty quantification, and engineering judgment.

In Section~\ref{sec8}, A. Scotti raises a foundational concern about the epistemic and educational pipeline: if AI systems build on accumulated knowledge, “who will provide the knowledge for the intelligence in the medium and long term?” This “burning question” reframes responsible AI as a long-horizon issue involving education, experimentation, mentorship, and the preservation of scientific and engineering culture.

In what follows, A. Cunha~Jr. contributes in three additional fronts. In Section~\ref{sec9}, he examines the implications of AI for engineering education, discussing the need to modernize curricula so that future mechanical engineers develop complementary competencies in engineering fundamentals, artificial intelligence and data-driven methods, ethical and professional responsibility, and human--agent teaming. Section~\ref{sec10} addresses ethics and accountability in engineering applications of AI, synthesizing cross-cutting risks—including over-reliance, automation bias, traceability gaps, and misuse of generated technical content—and formulating recommendations for responsible practice, professional governance, and curriculum modernization. In Section~\ref{sec11}, he discusses validation, uncertainty quantification, and certification, emphasizing that predictive performance alone is insufficient for engineering deployment and that trustworthy AI requires rigorous validation, quantified uncertainty, traceability, governance, and life-cycle monitoring.

Finally, in Section~\ref{sec12}, we integrate these threads into a compact set of principles and forward-looking recommendations for the responsible and constructive use of AI in ME, spanning industrial practice, university curricula, research directions, and the emerging paradigm of human--AI collaboration.

\begin{figure}[ht!]
\centering
\includegraphics[width=0.95\linewidth]{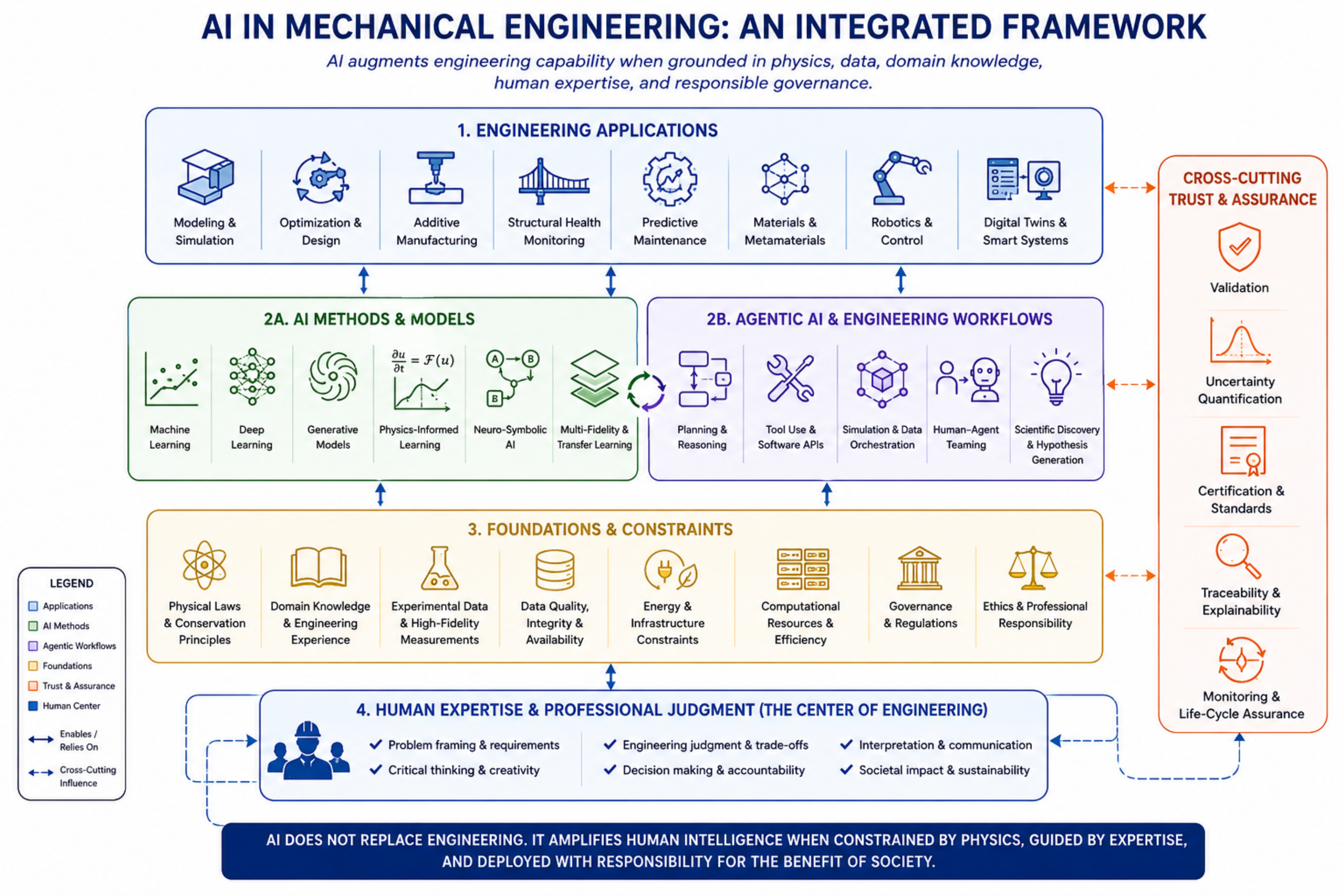}
\vspace{2mm}
\caption{System-level view of artificial intelligence in ME. Engineering applications are enabled by AI methodologies, including machine learning, physics-informed learning, generative models, and emerging agentic systems that coordinate simulations, optimization, and data-driven workflows. These capabilities remain grounded in physical laws, domain knowledge, experimental evidence, energy infrastructure, and governance frameworks. Validation, uncertainty quantification, certification, and traceability act as cross-cutting requirements across all layers, ensuring that AI systems can be deployed responsibly in engineering practice. The diagram emphasizes that AI functions as an embedded component of engineering systems rather than as an autonomous substitute for engineering judgment.}

\label{fig:AI_system_view}
\end{figure}

Figure~\ref{fig:AI_system_view} summarizes the system-level perspective adopted in this article, illustrating how AI methods interact with engineering applications while remaining constrained by physical laws, energy infrastructure, and governance requirements.

\section{How contemporary AI operates}
\label{sec2}

AI has transitioned in a relatively short time from a subject largely associated with science fiction to a pervasive technological presence in everyday life and professional practice. Today, AI systems are routinely embedded in applications such as biometric authentication, recommendation systems, virtual assistants, automated customer support, image analysis, autonomous robotics, and content generation. In engineering contexts, AI increasingly supports tasks ranging from data analysis and modeling to optimization and decision support \cite{Ciccone2023AI_AM_BSME,Abdelhamid2024ML_AM_BSME}, including applications in fluid dynamics \cite{Kutz2017,Silva2024}, weather forecasting \cite{Porto2022,Mouatadid2023}, seismic analysis \cite{Burkle2022,cunha2020p104344,Lacerda2021,Valentin2019p474}, as well as in advanced material design \cite{Bessa2017p633,Bessa2019,Chi2021p112739,Sanu2025}, and life-cycle simulation within integrated digital engineering frameworks such as digital twins \cite{McClellan2022p20210204,NAE2024DigitalTwins,Zhang2025p117863}. Despite this widespread adoption, fundamental questions remain regarding how AI systems operate, what they can reliably accomplish, and where their limitations lie \cite{Mitchell2023,RussellNorvig2021AIMA}.

At its core, AI comprises a set of statistical and computational techniques for identifying patterns in data and extracting actionable information \cite{JordanMitchell2015,NASEM2026Statistics}. By processing sufficiently large and representative datasets—such as images, text, sensor measurements, or behavioral records—AI systems learn statistical regularities that enable classification, prediction, and decision-making. In this sense, learning in AI is example-driven: performance improves through exposure to data, in a manner analogous to experience-based learning in humans, albeit through fundamentally different mechanisms.

Three broad learning paradigms underpin most modern AI systems. In supervised learning, models are trained on labeled data, in which correct outputs are provided during training. This paradigm is widely used in applications such as spam detection, fault classification, medical image analysis, and travel-time prediction. In unsupervised learning, data are unlabeled, and the system must identify intrinsic structures, clusters, or anomalies on its own. This approach is commonly applied to customer segmentation, pattern discovery in high-dimensional data, and anomaly detection in areas such as finance or industrial monitoring. Reinforcement learning, in contrast, is based on trial-and-error interaction with an environment, in which actions are evaluated by rewards or penalties \cite{SuttonBarto2018}. This paradigm is particularly relevant for control, robotics, autonomous vehicles, and sequential decision-making problems \cite{OConnell2022}.

Recent advances in foundation models and large language models (LLMs) have also enabled a new class of AI architectures known as agentic systems. Unlike traditional ML models, which typically perform a single prediction task, agentic systems combine language models with memory, planning, tool use, and iterative feedback mechanisms, allowing them to coordinate multi-step workflows toward a specified objective. The notion of an intelligent agent is not new in AI and traditionally refers to an entity capable of perceiving its environment, making decisions, and taking actions to achieve specified goals \cite{RussellNorvig2021AIMA,Wooldridge2009}. What is new is that modern LLMs now provide a flexible interface that enables agents to interact with software tools, databases, simulation environments, sensors, and external knowledge sources via natural-language instructions. This development has given rise to a rapidly growing research area focused on LLM-based agents capable of autonomous planning, tool use, and task execution \cite{Xi2025}.

In contrast to conventional ML models, which typically perform a single inference task, agentic systems operate through iterative decision loops involving reasoning, planning, memory, tool use, and feedback \cite{Xi2025,Wang2024AgentsSurvey,Plaat2026}. Rather than generating a single prediction or response, an agent can decompose a complex objective into multiple subtasks, invoke specialized computational tools, evaluate intermediate results, and adapt subsequent actions accordingly. In this sense, agentic systems can be viewed as orchestrators of computational workflows rather than standalone predictive models.

\begin{figure}[ht!]
\centering
\includegraphics[width=0.95\linewidth]{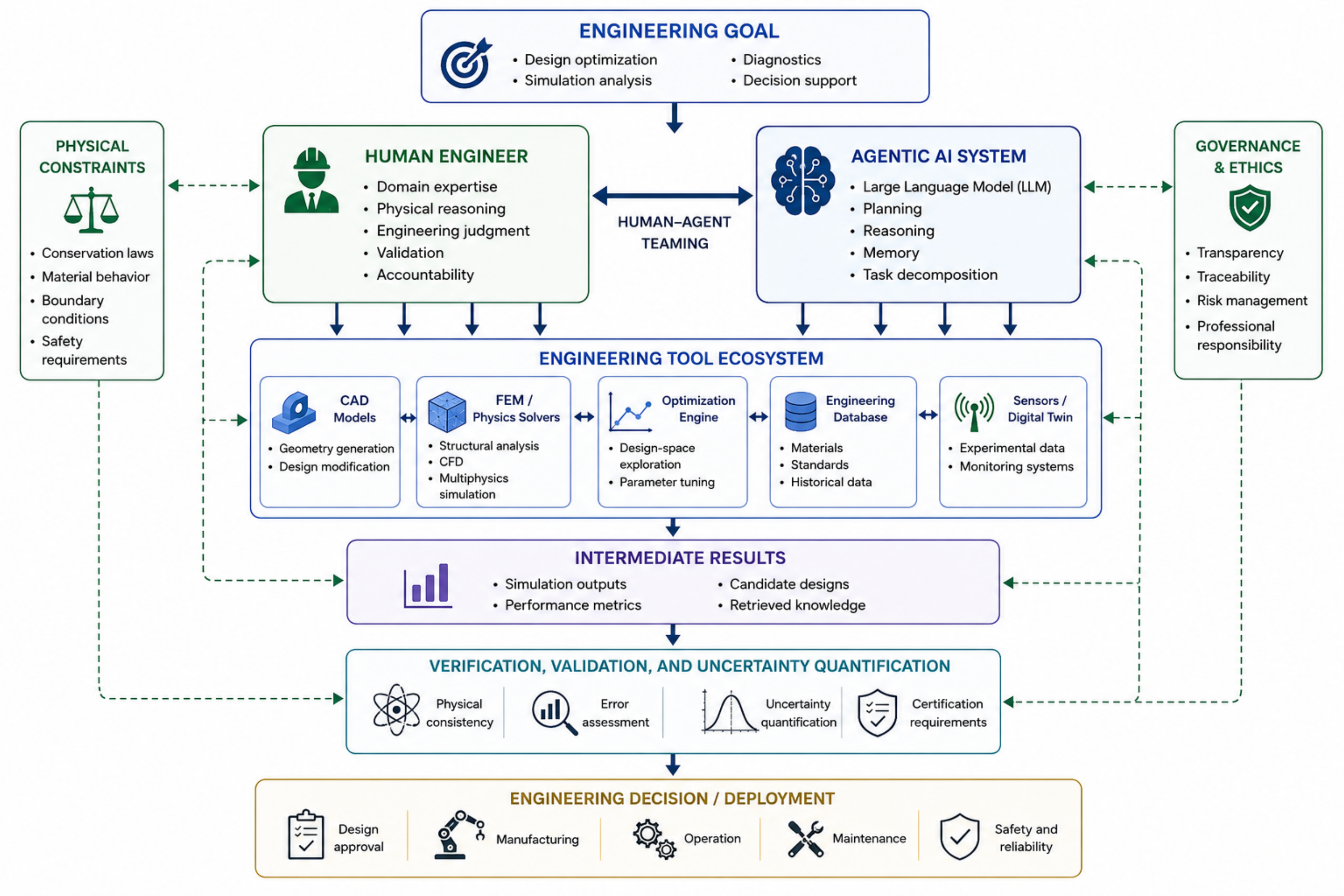}
\vspace{2mm}
\caption{Conceptual architecture of a human-agent engineering workflow. Engineering objectives are pursued through collaboration between a human engineer and an agentic AI system, which interact with an ecosystem of computational tools including CAD models, physics solvers, optimization engines, engineering databases, sensors, and digital twins. The agentic AI system assists with planning, reasoning, information retrieval, and workflow orchestration, while the human engineer provides domain expertise, physical reasoning, engineering judgment, validation, and accountability. Intermediate results generated throughout the workflow undergo verification, validation, and uncertainty quantification before supporting engineering decisions and deployment. Physical constraints, governance requirements, and ethical considerations act as cross-cutting influences across all stages. The diagram emphasizes that agentic AI can coordinate complex engineering workflows, but trustworthy deployment ultimately depends on human oversight, physical consistency, uncertainty awareness, and professional responsibility.}
\label{fig:agentic_ai_workflow}
\end{figure}

Figure~\ref{fig:agentic_ai_workflow} provides a conceptual representation of a human-agent engineering workflow, emphasizing that AI systems operate within a broader ecosystem of computational tools, physical constraints, validation procedures, and human oversight.

These capabilities are increasingly attracting attention in scientific computing and engineering. Emerging applications include autonomous scientific discovery, AI-assisted computer-aided engineering, self-driving laboratories, simulation-driven design, and scientific machine learning workflows \cite{DOEGenesis2025,Guo2026,Wang2023}. In ME, these capabilities may support automated simulation campaigns, design-space exploration, topology optimization, digital twins, structural health monitoring, and autonomous laboratory experimentation. While such systems can automate portions of complex engineering processes, their effectiveness remains fundamentally dependent on human-defined objectives, domain knowledge, physical models, and validation procedures.

The growing interest in agentic AI stems from its potential to automate portions of scientific and engineering workflows that traditionally require extensive human coordination. By combining reasoning, software interaction, and iterative execution, such systems may reduce the time required for simulation campaigns, design exploration, data analysis, and model development. Recent initiatives in scientific machine learning and autonomous engineering view agentic systems as a potential pathway toward accelerated discovery and productivity gains in science and engineering \cite{DOEGenesis2025,Guo2026,Jiang2026}. At the same time, the emergence of large-language-model-based agents has stimulated significant research efforts aimed at understanding how planning, memory, tool use, and multi-agent collaboration can be integrated into reliable autonomous workflows \cite{Xi2025,Wang2024AgentsSurvey}.

Many of these approaches rely on artificial neural networks, computational architectures inspired by biological neurons and organized in layers of interconnected units. During training, neural networks adjust internal parameters to perform specific tasks, such as image recognition, language translation, or signal classification. Deep neural networks, which involve many layers and parameters, have enabled remarkable advances in recent years \cite{Herrmann2025,GoodfellowBengioCourville2016,LeCun2015Nature,LinTegmarkRolnick2017}, but often require substantial computational resources, large datasets, and significant energy consumption \cite{Strubell2019,IEA_Energy_and_AI_2025}. The development and training of large-scale models—particularly generative models for text, images, or audio—may involve specialized hardware, extensive infrastructure, and considerable economic and environmental costs.

The rapid pace of technical improvement is evident in recent benchmark trajectories: according to the Stanford AI Index 2025, performance increased sharply within a single year on demanding evaluations such as MMMU, GPQA, and SWE-bench, reflecting the speed with which modern AI systems are advancing on complex multimodal, scientific, and software-engineering tasks \cite{stanford_ai_index_2025}.

Another important characteristic of most current AI systems is task specificity. Models are typically optimized for narrowly defined problems and perform poorly outside their intended scope \cite{Abdelhamid2024ML_AM_BSME}. An AI trained for image classification, for example, cannot directly process audio signals or engage in natural-language dialogue. While recent multimodal models can handle multiple input types, such generality remains limited, and the development of truly versatile, general-purpose AI systems remains an open research challenge \cite{Bommasani2022,Zhao2023Survey}.

This rapid progress, however, should not be confused with robust general intelligence. Assessments published in 2025--2026 note that leading general-purpose AI systems now perform strongly in domains such as coding, mathematics, and scientific reasoning, yet remain unreliable in real-world settings and continue to struggle with complex planning, logically structured problem solving, and other tasks that demand stable reasoning under uncertainty \cite{InternationalAISafetyReport2026,stanford_ai_index_2025}.

From a functional standpoint, contemporary AI systems can be broadly categorized into predictive models, generative models, and increasingly, agentic systems that combine prediction, generation, planning, tool use, and action within a common workflow \cite{Bommasani2022,Plaat2026,Wang2024AgentsSurvey}. By coordinating multiple computational tasks and external resources, agentic systems can support complex engineering workflows involving simulation, optimization, data analysis, and decision support. It is important to interpret these capabilities operationally rather than cognitively. Terms such as planning, tool orchestration, memory mechanisms, and iterative decision processes describe algorithmic processes that emerge from statistical inference, tool orchestration, and iterative execution, rather than from human-like understanding or consciousness. In engineering applications, anthropomorphic interpretations of AI systems may lead to misplaced trust and unrealistic expectations regarding reliability, autonomy, and accountability. Despite their impressive capabilities, neither class of system possesses understanding, intention, or consciousness in a human sense. Terms such as ``learning,'' ``knowing,'' or ``thinking'' are best interpreted as convenient metaphors for statistical optimization processes rather than indicators of genuine cognition \cite{Bender2021,Mitchell2023}.

This distinction is particularly important in engineering contexts, where reliability, interpretability, accountability, and physical consistency are critical \cite{Haywood-Alexander_Liu_Bacsa_Lai_Chatzi_2024}. AI systems do not reason or comprehend in the way engineers do; they operate within the limits imposed by their data, architectures, and training objectives. A clear understanding of these foundations and limitations is therefore essential for the responsible and constructive integration of AI into ME research, education, and professional practice.

\section{Industrial adoption of AI and workforce transformation}
\label{sec3}

This section aims to present a perspective on how industries currently perceive and use AI, particularly generative AI, and how organizations are positioning themselves in the early stages of adoption. In recent years, several technologies have emerged—such as robotics and the Internet of Things (IoT)—and are accompanied by expectations of profound and definitive transformations across industries, production processes, occupations, and jobs. However, these changes occurred more gradually and with less disruption than initially anticipated. With the emergence of generative AI, similar questions have resurfaced. Unlike previous technologies, however, there are strong indications that generative AI may drive transformations at multiple levels, making its impact increasingly difficult to ignore.

This perception is shared by industry and is reinforced by the study "The State of AI: How Organizations Are Rewiring to Capture Value", published in March 2025 by QuantumBlack AI (McKinsey) \cite{McKinsey_State_of_AI_2025}. The study highlights that organizations are beginning to implement structural changes to capture future value from generative AI. Large companies are leading this movement by redesigning workflows and assigning senior leaders to strategic roles, including AI governance responsibilities. The study also notes that 28\% of respondents report that their CEO is responsible for overseeing AI governance, underscoring both the strategic importance of the topic and that AI-driven transformation is being led at the highest executive level.

In addition, AI adoption has increased significantly over the past three years. The use of AI in at least one business process increased from 55\% in 2023 to 78\% in 2025, a trend broadly consistent with other global tracking efforts \cite{McKinsey_State_of_AI_2025,stanford_ai_index_2025}. The growth of generative AI usage has been even more pronounced, increasing from 33\% to 71\% over the same period \cite{McKinsey_State_of_AI_2025}. Organizations report tangible benefits in the processes where AI is applied, suggesting that adoption will continue to intensify. Nevertheless, current use cases remain concentrated in a limited set of applications: 63\% of organizations use AI for text generation, 36\% for image generation, 27\% for code creation, 13\% for video, voice, and music, and only 2\% for other uses. This indicates that, although benefits are already being realized, the overall impact remains marginal and is not yet truly transformative \cite{McKinsey_State_of_AI_2025}.

Independent analyses of enterprise AI deployment highlight a significant gap between adoption and measurable organizational transformation. A recent study from MIT NANDA describes a “GenAI Divide,” in which widespread experimentation with generative AI tools contrasts with limited operational impact: although most organizations have piloted AI systems, only a small fraction report measurable economic value from their deployment \cite{MIT_State_of_AI_Business_2025}. These findings suggest that successful AI integration depends less on access to models than on the ability to adapt workflows, incorporate domain knowledge, and build systems that learn from real operational feedback.

This limited impact is largely due to the early stage of adoption. As industries gain greater mastery of the technology and integrate AI more deeply with scientific models, data, and simulation tools, broader and more transformative applications are expected to emerge \cite{Brunton2021AIAA}. In this context, AI is increasingly viewed as an enabler of engineering activities rather than a replacement for established methods and expertise.

Although large foundation models have attracted substantial attention, many industrial AI deployments occur at the edge, close to physical assets and operational processes. Embedded AI systems are increasingly used in smart sensors, industrial monitoring, condition assessment, predictive maintenance, and control applications, where latency, energy efficiency, reliability, and limited computational resources often take precedence over raw model size \cite{Banbury2021,Hamasha2025}.

Finally, the study indicates that AI is reshaping the skills required within the workforce. While these skills remain relatively scarce, they are now more available than in the past. This reality highlights the need for new educational methodologies, particularly within universities, which must update curricula to better align with evolving industrial demands. Beyond digital and technical competencies—such as AI and data science, which should become mandatory—future professionals must also develop skills that AI cannot replicate. These include creativity, collaboration, leadership, emotional intelligence, critical thinking, problem-solving, sustainability awareness, and ethical judgment \cite{NASEM2025AIFutureWork}.

In summary, industries clearly recognize the strategic importance of AI and the transformations it is expected to bring to society and business operations. While practical applications remain in their early stages and are largely focused on text, image, and code generation, broader adoption is anticipated. Preparing the future workforce for this new reality is therefore critical, and universities and academia play a central role in this effort.

\section{Energy challenges and opportunities for {AI}}
\label{sec4}

The rapid expansion of AI is creating an increasingly visible and nontrivial challenge: the energy demand associated with large-scale data centers. As the physical backbone of AI development and deployment, data centers concentrate massive computational workloads within confined spaces, resulting in very high electricity consumption and power densities. Recent assessments indicate that the growth of AI workloads is a material driver of electricity demand, with data centers emerging among the fastest-growing loads in several power systems worldwide \cite{IEA2025EnergyDemandFromAI,ShehabiEtAl2024USDataCenters,Xiao2025}. This infrastructural pressure is closely linked to the scaling behavior of frontier AI systems. In fact, the training of notable AI models has been doubling on a timescale of roughly five months, while training power requirements continue to rise on an annual basis \cite{stanford_ai_index_2025}. Global analyses of AI governance emphasize that compute capacity, data infrastructure, and energy availability are rapidly becoming critical determinants of national technological capability. Current studies suggest that a small number of countries concentrate most large-scale AI data centers and computing resources, while many regions remain largely without access to such infrastructure \cite{OxfordAIGovernance2025}. 

\begin{figure}[ht!]
\centering
\includegraphics[width=0.95\linewidth]{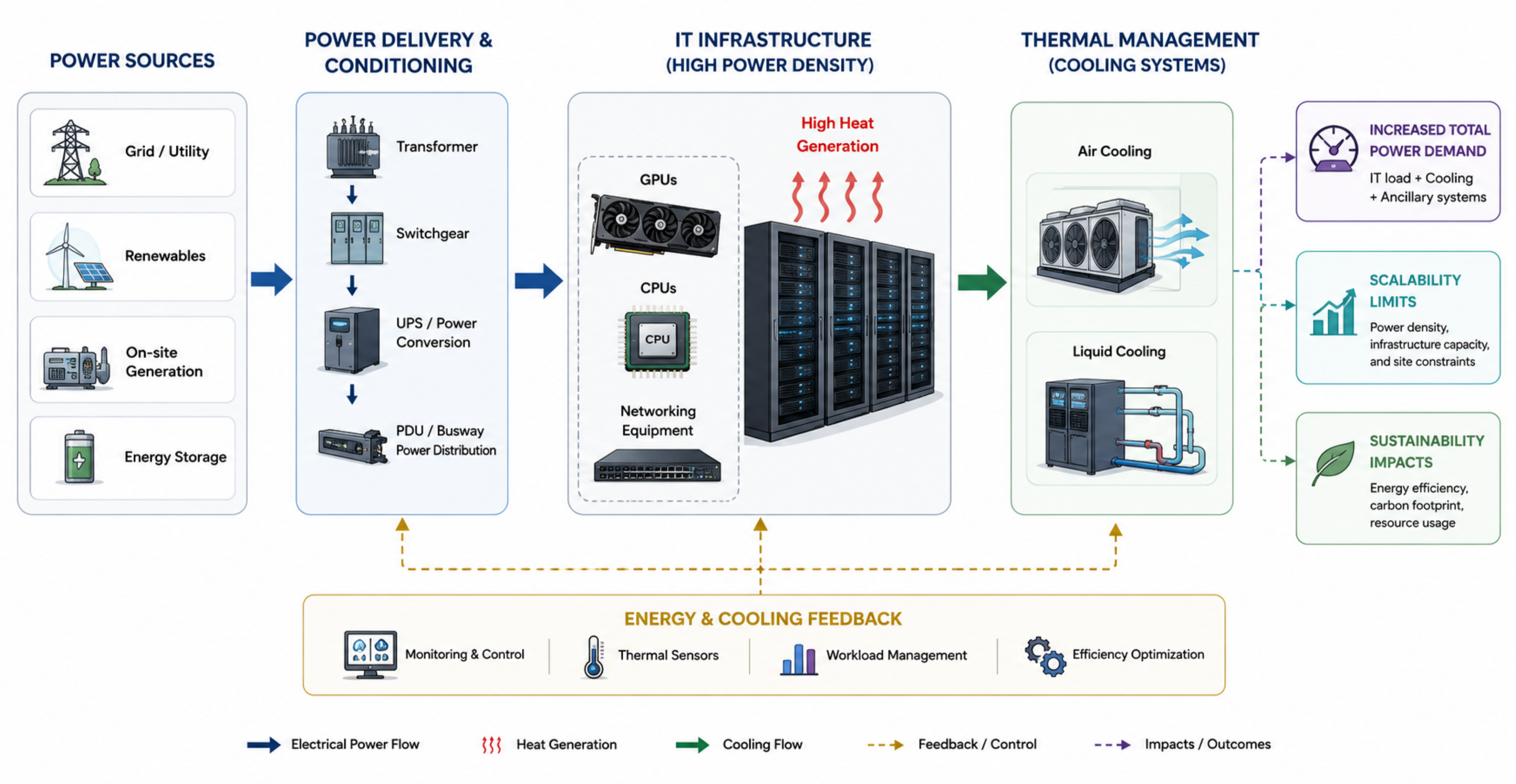}
\vspace{2mm}
\caption{Schematic representation of the energy flow and infrastructure requirements of AI-driven data centers. Electrical power from primary sources is delivered through distribution and conditioning systems to high-density computing hardware (GPUs, CPUs, and networking equipment), which generate substantial heat loads. Thermal management subsystems—including liquid- and air-cooling—require additional electrical power, thereby increasing total facility power demand and shaping the scalability and sustainability of AI infrastructure.}
\label{fig:Energy_AI_DC}
\end{figure}

The principal components contributing to energy demand in AI-oriented data centers are schematically illustrated in Figure~\ref{fig:Energy_AI_DC}, highlighting the interaction between IT load, power conditioning, and thermal management.

The energy intensity of AI-driven data centers arises from multiple factors. First, modern AI training and inference rely heavily on high-performance graphics processing units (GPUs), such as those produced by Nvidia, which operate at high utilization levels and require continuous, reliable power supply. Second, the resulting heat dissipation necessitates sophisticated cooling systems, including chilled-water plants, evaporative cooling, and, increasingly, direct liquid cooling. In many facilities, cooling and power-conditioning infrastructure can account for a substantial fraction of total electricity consumption, making thermal management a central engineering challenge \cite{ShehabiEtAl2024USDataCenters}. As a consequence, data centers exhibit not only high total electricity demand but also stringent requirements for power quality, redundancy, and availability. 

Projections suggest that, if current trends persist, data centers could account for a significant share of electricity consumption in major economies within the next decade. In the United States, for example, recent analyses indicate a sharp increase from approximately 4.4\% of total electricity demand in 2023 to 12\% toward the end of the decade, driven largely by AI-related workloads \cite{ShehabiEtAl2024USDataCenters,IEA2025EnergyDemandFromAI}. This rapid escalation has intensified concerns regarding grid capacity, carbon emissions, water use for cooling, and the long-term sustainability of AI infrastructure. These concerns are reinforced by recent estimates of training-related carbon emissions, which have grown sharply with model scale, highlighting that the environmental footprint of AI depends not only on where electricity comes from, but also on the computational intensity and efficiency of the systems being developed \cite{stanford_ai_index_2025}.

In response, the technology sector is actively exploring alternative and complementary energy solutions. These include greater reliance on renewable generation, long-term power purchase agreements, advanced cooling technologies, energy storage, small modular nuclear reactors, geothermal energy, and hydrogen-based systems \cite{Alkrush2024p246}. Hydrogen-powered data centers, in particular, have attracted attention as a means of decoupling data-center deployment from grid constraints and enabling faster siting and scaling \cite{HydrogenEurope2025DataCentres}. Industry initiatives suggest that hydrogen fuel cells could provide both primary and backup power, offering high reliability and potentially lower life-cycle emissions when supplied with low-carbon hydrogen.

At present, however, most hydrogen production remains dominated by fossil-based routes, particularly steam reforming of methane, which is carbon-intensive \cite{IEA2023GlobalHydrogenReview}. Water electrolysis (WE) is a cleaner alternative when powered by low-carbon electricity, but it is constrained by low efficiency (typically below 50\%) and high capital and operating costs \cite{Buttler2018Electrolysis}. These limitations constrain the large-scale deployment of green hydrogen for energy-intensive applications such as data centers.

In this context, alternative hydrogen production pathways are being investigated. One promising approach involves hydrogen generation from the reaction of water with aluminum, known as the Alkaline Aqueous Corrosion of Aluminum (AACA) process. This method generates hydrogen without external electrical input, produces no direct carbon emissions, and can enable cogeneration due to its exothermic nature. Moreover, aluminum is the third most abundant element in the Earth’s crust, and scrap aluminum can be fully recycled. Recent experimental and system-level studies suggest that aluminum-based hydrogen generation can be coupled with fuel cells to deliver competitive power output for stationary and mobile applications \cite{Polla2024IJHE,Guerra2025TopCat}.

Not all engineering applications of AI rely on large-scale cloud infrastructure. Many ME systems employ edge AI, in which inference is performed directly on embedded processors, industrial controllers, smart sensors, or cyber-physical devices. These applications operate under constraints that differ markedly from those of data centers, emphasizing low power consumption, limited memory, real-time operation, and hardware efficiency \cite{Banbury2021}. Consequently, energy-aware AI must be considered across a spectrum ranging from hyperscale computing facilities to resource-constrained embedded systems.

From a broader perspective, it is increasingly clear that no single energy solution is likely to dominate the future of AI infrastructure. Instead, a portfolio of approaches—combining grid-based renewables, advanced cooling, energy efficiency measures, storage, hydrogen in various forms, and other low-carbon energy sources—will likely be adopted, depending on regional conditions, climate, grid characteristics, and regulatory frameworks. For Brazil, this challenge presents both risks and opportunities. The country benefits from a relatively low-carbon electricity mix dominated by hydropower, growing penetration of wind and solar energy, and favorable climatic and geographic conditions for renewable generation \cite{EPE2025BrazilEnergy}. All the same, high ambient temperatures in many regions increase cooling demand, and regional disparities in grid capacity may influence data center siting. These factors position Brazil as a potentially attractive location for AI infrastructure, provided that energy planning, thermal management, and sustainability considerations are addressed in an integrated manner.

These opportunities are increasingly recognized at the national level. The PBIA 2024--2028 \cite{PBIA2024} identifies advanced computational infrastructure as a strategic asset for scientific and technological development, including investments in high-performance computing facilities and associated digital infrastructure. Such initiatives directly connect AI development with the energy, cooling, and infrastructure challenges discussed in this section.

A similar strategic direction can be observed in Europe through the EU AI Continent Action Plan, which promotes major investments in AI factories, high-performance computing infrastructure, sustainable data centers, and advanced cloud capabilities to support the next generation of AI systems. Together, these initiatives illustrate how computational infrastructure and energy availability are increasingly recognized as strategic assets for scientific, technological, and industrial competitiveness \cite{EUAIContinent2025}.

Ultimately, addressing the energy demands of AI will require close collaboration between mechanical engineers, energy-system planners, data-center designers, and policymakers. Rather than seeking a single technological remedy, responsible AI deployment calls for system-level engineering solutions that balance performance, reliability, environmental impact, and regional realities.

\section{Physical knowledge and reasoning for better AI systems in engineering}
\label{sec5}

Taken together, the previous sections highlight a central tension: AI is being rapidly adopted in engineering workflows, yet many high-consequence ME tasks cannot rely solely on data-driven pattern recognition. Data scarcity, distribution shifts, and the need for traceability and physical consistency motivate approaches that combine learning with structured representations of domain knowledge and governing principles. This motivates the present section, which discusses neuro-symbolic and physics-informed approaches to improving generalization, interpretability, and trust when AI is deployed in safety-relevant ME contexts \cite{karniadakis2021physicsml,raissi2019pinns}.

Neuro-symbolic artificial intelligence (NSAI) combines the pattern-recognition capabilities of neural networks with the logic-based inference of symbolic reasoning, and has shown significant promise across several domains, including ME \cite{Bhuyan2024NSAI}. By integrating learning from data with structured reasoning, NSAI offers a potential pathway to overcome some of the limitations of purely data-driven approaches. However, important challenges remain before such approaches can be reliably deployed in engineering contexts.

Mechanical systems are governed by complex physical laws and constraints that are often difficult to encode symbolically or to infer directly from data alone. Integrating physical models with neural architectures in a coherent and computationally efficient manner, therefore, remains a major challenge. In addition, data sparsity and noise—common in many ME applications such as structural health monitoring or fluid dynamics—limit the effectiveness of learning approaches that rely exclusively on large, high-quality datasets. In this context, NSAI must bridge the gap between data and domain knowledge by explicitly incorporating physical principles into the learning process. 

Furthermore, real-time reasoning is essential in applications such as robotics, control systems, and adaptive manufacturing, yet current neuro-symbolic models often lack the speed and robustness required for such tasks. Explainability represents an additional critical requirement, as engineering applications frequently involve safety, certification, and regulatory compliance, while many existing AI models still operate as black boxes \cite{Rudin2019Interpretable}.

An additional, increasingly relevant implication of these hybrid approaches concerns computational and energy efficiency. By constraining learning with physical laws, symbolic structures, and domain knowledge, neuro-symbolic and physics-informed methods can significantly reduce the size of the hypothesis space explored during training. This more directed and informed learning process mitigates the need for large-scale brute-force pattern search, extensive retraining cycles, and oversized models, which are characteristic of purely data-driven approaches. As a result, such methodologies have the potential to lower computational cost and energy consumption, particularly in engineering applications where data are scarce, models must generalize reliably, and training or inference occurs under resource constraints.

In light of the current state of the art, several research directions merit particular attention. One important avenue is the development of physics-informed neuro-symbolic models in which governing equations and physical constraints are embedded directly within the learning process, with the aim of improving generalization and reducing dependence on large datasets. Another promising direction is the development of domain-specific symbolic ontologies that guide learning and reasoning in ME contexts, such as thermodynamics, kinematics, and material behavior. Research should also explore modular architectures that explicitly separate perception components, handled by neural networks, from reasoning components, handled by symbolic systems, thereby enabling more interpretable, adaptable, and maintainable solutions \cite{AlHajAli_JII_2025}. 

In parallel, the development of real-time inference frameworks is required to support applications in autonomous systems, smart manufacturing, and adaptive control. Finally, benchmarking and standardization of NSAI tools in ME contexts would help accelerate adoption and build trust in their use. Enhancing cognition in neuro-symbolic systems through causal reasoning, contextual awareness, and goal-directed planning is also seen as a key requirement for future engineering applications \cite{AlHajAli_AnnualReviewsInControl_2024}. 

These challenges motivate a closer examination of how learning-based methods can be integrated with physical constraints in practical model development, which is the focus of the next section.


\section{Learning-based model development under physical constraints}
\label{sec6}

Building on the broader discussion of hybrid and physics-integrated AI, this section focuses on model development and diagnostics in mechanical systems, where inverse and often ill-posed inference problems make physical structure especially important.

Model development for mechanical systems (and related ones such as electromechanical and biomedical) is the process of constructing mathematical representations that relate system structure, parameters, and dynamics to observable behavior. Diagnostics is traditionally treated as a distinct activity concerned with fault detection, isolation, and identification. We adopt a unified view in which diagnostics is a specialized instance of model development under partial observability, uncertainty, and structural constraints. Model development generally relies on indirect measurements, such as vibration, current, voltage, acoustic emission, or thermal response, to infer hidden system properties, including parameters, constitutive behavior, and degradation. The resulting inference problem is inverse in nature and is often ill-posed in the classical sense described by Hadamard, lacking uniqueness or stability of the inverse mapping \cite{Hadamard1923,Tikhonov1977}. While this ill-posedness motivates regularization and prior structure, it does not preclude meaningful inference when physics and data are combined appropriately.

Recent advances in machine learning (ML) have created new opportunities for model development and diagnostics by enabling flexible approximation of complex, nonlinear relationships from data, including operator-learning approaches for accelerating simulations \cite{Azizzadenesheli2024,Lu2021} or emerging architectures designed to improve efficiency and accuracy in scientific computing \cite{Bizzi2025,Liu2024KAN,Park2025}. However, the effectiveness and trustworthiness of such models depend critically on how learning is integrated with physical knowledge. Machine learning should ideally be viewed as an enabling component within physics-informed model development \cite{Brunton-Kutz:2022}, rather than as a replacement for mechanistic understanding. It offers powerful tools for approximating unknown mappings, discovering latent representations, and assimilating large volumes of heterogeneous data. In model development, data-based approaches can approximate complex constitutive behavior, nonlinear dissipation, contact phenomena, or parameter drift that are difficult to model from first principles alone. In diagnostics, ML enables automated feature extraction, fault classification, and probabilistic inference from high-dimensional sensor data.

Importantly, ML does not attempt to compute a physical inverse mapping. Instead, it constructs statistically regularized surrogates that remain stable under perturbations to the data distribution encountered during training \cite{Bishop2006,Freno2019p250}. These surrogates can be highly effective \emph{within their domain of validity} \cite{Trehan2017}, providing practical solutions to inference problems that are otherwise ill-posed in a classical sense.
However, purely data-driven models derive their stability from empirical regularization rather than physical identifiability. As a result, their reliability depends on the ``completeness'' of the data distribution and the relevance of learned correlations to the underlying dynamics.

The integration of physics into ML–based model development is hence essential for robustness, interpretability, and trustworthiness. Physical laws, constitutive relations, and system dynamics impose constraints that restrict the space of admissible models to those that are physically realizable. These constraints mitigate non-uniqueness and instability in inference by ruling out models that fit data but violate known principles such as conservation of energy and mass, causality, compatibility conditions, etc. \cite{Abbasi2023}. From a robustness perspective, physics-informed models are less sensitive to noise and spurious correlations. By enforcing consistency with governing equations, physics acts as a stabilizing structure that complements statistical regularization. This is particularly important for extrapolation across operating regimes, loading conditions, or environmental variability where data-driven models alone consistently underperform \cite{Kappaganthu2011}.

Physics also provides semantic grounding for model parameters and latent variables. In model development, this grounding enables interpretability, validation, and model reuse across systems and applications. Nonlinear physics, in particular, is critical for models to capture real phenomena \cite{Nataraj2025}.  In diagnostics, it supports causal reasoning \cite{PearlMackenzie2018BookOfWhy} and facilitates human trust in automated decision-making, especially in safety-critical contexts such as structural health monitoring, rotating machinery  \cite{Kappaganthu2011,Laflamme2026SHM,Mohamad2023}, electromechanical systems \cite{Doebling1998,Ljung1999}, biomedical applications \cite{Borges2025,Nataraj2025a}, as well as reacting flows and combustion systems \cite{Shead2023}.

Hybrid physics-informed learning frameworks naturally emerge from this perspective. In such approaches, physics-based models provide the structural backbone, while ML captures residual effects, unmodeled dynamics, or uncertainty \cite{Ezvan2026,raissi2019pinns,Liu2025,Abbasi2023,Abbasi2023a}. This division of roles preserves physical coherence while leveraging the expressive power of data-driven methods. For digital twins and long-horizon predictive models for dynamic systems, reliance on purely data-driven learning is insufficient. Sustained predictive fidelity requires physics-informed structure to ensure stability, interpretability, and meaningful uncertainty quantification (UQ) \cite{Abbasi2022,MLUQ2025p118061}. When embedded within such a structure, ML becomes a powerful enabler rather than a source of fragility.

An important extension of hybrid modeling is the use of multi-fidelity frameworks, in which information from simulations, experiments, reduced-order models, and machine-learning surrogates is combined across different levels of accuracy and computational cost. Multi-fidelity methodologies can improve predictive performance, reduce data requirements, and support UQ by leveraging complementary information sources \cite{Yi2025,Trehan2017,Freno2019p250,Willcox2018,Zhang_2024}.

A related challenge concerns the so-called Sim2Real gap. Many AI models are trained using numerical simulations or synthetic datasets but are ultimately deployed in physical systems whose operating conditions, noise characteristics, and uncertainties differ from those represented in the training data. Bridging this gap is essential for reliable deployment of digital twins, predictive maintenance systems, autonomous engineering workflows, and other AI-enabled engineering applications. Physics-informed learning, experimental validation, UQ, and continuous model updating are increasingly recognized as important mechanisms to improve transferability from simulation to reality \cite{Katyara2025,Li2025}.

In summary, while AI provides new opportunities to address complexity and data richness, physics remains essential for constraining inference, enabling extrapolation, and establishing trust. Physics-informed machine learning provides a rational pathway to combine data-driven flexibility with mechanistic understanding, thereby supporting robust and interpretable models for complex mechanical systems.

\section{Representative mechanical engineering applications}
\label{sec7}

The preceding sections discussed AI from a methodological perspective, emphasizing the integration of data-driven learning with physical knowledge and engineering constraints. To illustrate how these ideas are being translated into practice, this section highlights representative application areas within ME where AI is already influencing research and industrial workflows. The objective is not to provide a comprehensive review of AI applications in ME, but rather to illustrate how the principles discussed in Sections~\ref{sec5} and \ref{sec6} manifest in representative domains where physics-informed and hybrid AI approaches are already influencing engineering practice.

The selected examples span multiple scales and functions within ME. They include manufacturing processes (additive manufacturing), structural integrity assessment (structural health monitoring, fatigue, and fracture), operational reliability (tribology and predictive maintenance), computational mechanics and scientific computing, dynamic-system modeling and control (digital twins and robotics), materials and structural design (multiscale materials engineering and topology optimization), and emerging AI-enabled scientific workflows. Together, these domains illustrate the breadth of AI adoption across the ME life cycle, from material and component design to operation, maintenance, and scientific discovery. Other important areas not discussed in detail include biomechanics, thermal systems, aerospace engineering, renewable-energy systems, and computational fluid dynamics, all of which are experiencing rapid adoption of AI-based methodologies.

\subsection{Additive manufacturing}

Additive manufacturing has emerged as one of the most active areas at the intersection of AI and ME. Modern manufacturing processes generate large volumes of heterogeneous data from cameras, thermal imaging systems, acoustic sensors, melt-pool monitoring devices, and machine-control systems. These data create opportunities for ML methods to support process monitoring, defect detection, quality assurance, process optimization, and digital twin development throughout the manufacturing life cycle \cite{Ciccone2023AI_AM_BSME,Abdelhamid2024ML_AM_BSME,Meng2020p2363}.

Particularly important are applications involving real-time monitoring of melt-pool dynamics, in-situ defect detection, and the characterization of process--structure--property relationships. In these settings, the objective is not merely to identify correlations in data, but to infer underlying physical mechanisms governing thermal transport, phase transformations, residual stress formation, and microstructural evolution. Consequently, many of the challenges discussed in Sections~\ref{sec5} and~\ref{sec6} naturally arise, including limited data availability, UQ, inverse modeling, and the need to combine ML with physical constraints. Recent advances in surrogate modeling, neural operators, and hybrid simulation frameworks suggest that AI may substantially accelerate process optimization and design exploration while preserving consistency with the governing physics \cite{Park2025,Azizzadenesheli2024}.

\subsection{Structural health monitoring, fatigue, and fracture}

Structural health monitoring, fatigue assessment, and fracture mechanics provide canonical examples of inverse problems in ME. In these applications, engineers seek to infer hidden damage states, crack initiation, crack propagation, material degradation, or remaining useful life from indirect observations such as vibration signals, strain measurements, acoustic emissions, thermal fields, and operational data. Because the underlying damage processes are only partially observable, the resulting inference problems are often ill-posed and affected by measurement uncertainty, environmental variability, and model inadequacy, among other factors.

Artificial intelligence has demonstrated significant potential in automated feature extraction, anomaly detection, fault classification, and prognostics. However, reliable deployment in safety-critical applications requires more than predictive accuracy. Fatigue and fracture processes are governed by physical mechanisms that must remain consistent with conservation laws, constitutive behavior, damage evolution, and structural dynamics. Consequently, structural health monitoring has become an important testbed for hybrid digital twins, uncertainty-aware ML, and physics-informed AI frameworks that combine data-driven inference with mechanistic understanding \cite{Laflamme2026SHM,Kappaganthu2011,Mohamad2023,Worden2018p139,WordenFarrar2007}. These applications exemplify many of the challenges discussed in Section~\ref{sec6}, including partial observability, non-uniqueness of inverse solutions, UQ, and the need to balance statistical learning with physical interpretability. In many safety-critical applications, explainability and traceability are also required to support certification, inspection, and engineering decision-making.

Recent studies have also explored AI-assisted crack detection, fatigue-life prediction, and damage-progression modeling \cite{Agrawal2019p779,Allen2022}, in which ML methods are combined with fracture mechanics principles to improve prognostic capability under variable loading conditions. Recent efforts increasingly incorporate UQ to assess predictive confidence, support risk-informed maintenance decisions, and guide inspection planning under uncertain operating conditions \cite{MLUQ2025p118061,Yi2025}.

\subsection{Tribology and predictive maintenance}

Tribological systems are central to ME because friction, wear, lubrication, and surface interactions strongly influence the performance, efficiency, reliability, and lifetime of engineered components. Bearings, gears, seals, turbines, compressors, and other rotating machinery continuously generate operational data through vibration sensors, acoustic measurements, temperature monitoring, oil analysis, and supervisory control systems. These data provide opportunities for AI to support condition monitoring, fault diagnosis, degradation assessment, and predictive maintenance \cite{Lei2018p799,Zonta2020p106889}.

Many tribological phenomena involve complex multi-scale interactions among material properties, contact mechanics, lubrication regimes, surface roughness, and operating conditions. As a result, purely empirical approaches often struggle to generalize beyond the conditions represented in the training data. Hybrid methodologies that combine physical models of wear, lubrication, and system dynamics with ML techniques offer a promising path toward more reliable prognostics and maintenance planning. As with the fatigue and structural health monitoring problems discussed previously, tribological applications require integrating UQ, physical consistency, and engineering judgment to transform data into actionable maintenance decisions. These systems therefore represent another important domain in which AI serves not as a substitute for engineering knowledge, but as a tool for enhancing diagnosis, prognosis, and operational decision support.

\subsection{Dynamics, control, robotics, and digital twins}

Mechanical systems are inherently dynamic, evolving over time under the influence of external forcing, control actions, environmental variability, and internal nonlinearities. Consequently, model development, system identification, state estimation, control, and diagnostics occupy a central role in ME. Recent advances in artificial intelligence have expanded the range of available tools for addressing these challenges, enabling data-driven system identification, adaptive control, anomaly detection, reduced-order modeling, and real-time decision support in complex dynamical systems \cite{Brunton-Kutz:2022,Brunton2021AIAA}.

The convergence of AI, sensing technologies, and computational modeling has also accelerated the development of digital twins \cite{Fuller2020p108952,Kapteyn2020DT,Ritto20211p07614,Silva2025DigitalTwin}, which integrate physical systems with continuously updated computational representations. In this context, ML can assist with state estimation, model updating, uncertainty characterization, and prediction under changing operating conditions. Similar ideas are increasingly appearing in autonomous systems and robotics, where learning-based methods are combined with physical models to support navigation, adaptation, and control in uncertain environments \cite{SuttonBarto2018,OConnell2022}. These applications highlight a recurring theme throughout this article: reliable performance requires not only learning from data but also incorporating physical constraints, UQ, and engineering knowledge. As digital twins become more sophisticated and autonomous systems become more capable, the challenge is increasingly shifting from data acquisition to the disciplined integration of learning algorithms with mechanistic understanding. A persistent challenge across many of these applications is the Sim2Real gap, in which models trained predominantly on simulation data may exhibit degraded performance when exposed to real operating conditions. Bridging this gap remains an active research area involving domain adaptation, transfer learning, hybrid modeling, and experimental validation.

\subsection{Computational mechanics and scientific computing}

Computational mechanics has become one of the principal domains where AI and ME intersect. Modern engineering analysis relies heavily on numerical models based on partial differential equations, including finite element, finite volume, boundary element, and meshfree formulations. These models enable prediction of structural, thermal, fluid, and multiphysics phenomena but often require substantial computational resources, particularly for large-scale simulations, optimization studies, UQ and propagation, and multiscale analyses.

Artificial intelligence is increasingly being used to augment traditional computational mechanics workflows through surrogate modeling, reduced-order modeling, operator learning, adaptive sampling, and data-driven acceleration of numerical solvers. Rather than replacing established numerical methods, these approaches seek to reduce computational cost while preserving predictive fidelity. Recent developments in neural operators and ML-based surrogate models have demonstrated the potential to approximate solution operators of complex partial differential equations, enabling rapid evaluation of parametric systems that would otherwise require repeated high-fidelity simulations \cite{Trehan2017,Freno2019p250,Azizzadenesheli2024,Park2025,Willcox2018}.

AI is also contributing to model reduction, error estimation, mesh adaptation, UQ and propagation, and real-time simulation environments. In many applications, ML is used to learn corrections to existing numerical models, quantify approximation errors, or construct efficient emulators for design optimization and digital twin applications. Such developments are particularly relevant in engineering settings where computational budgets, time constraints, or the need for repeated simulations make direct use of high-fidelity models impractical. When surrogate models replace expensive numerical simulations, UQ becomes essential for assessing model-form error, predictive credibility, and decision robustness. Recent developments combine machine learning with probabilistic and multi-fidelity methodologies to propagate uncertainty and support engineering decision-making under limited computational budgets \cite{Trehan2017,Freno2019p250,MLUQ2025p118061,Willcox2018,Yi2025,Zhang_2024}.

Another emerging direction is equation discovery, in which sparse regression, symbolic regression, and scientific machine learning techniques are used to infer governing equations directly from data while preserving physical interpretability \cite{Brunton2016}.

Despite these advances, computational mechanics remains fundamentally grounded in physical principles, mathematical modeling, and numerical analysis. The most successful AI-enabled approaches typically combine data-driven learning with governing equations, conservation laws, variational formulations, and UQ. This convergence is gradually transforming scientific computing from a purely simulation-driven paradigm toward an integrated framework in which physics-based models and machine learning cooperate to accelerate analysis, improve prediction, and expand the scope of computational engineering.

\subsection{Materials design, multiscale modeling, and topology optimization}

Artificial intelligence is increasingly transforming the way materials and structures are designed, analyzed, and optimized. Traditional materials engineering relies heavily on multiscale computational models that connect microscale features---such as grain morphology, porosity, defects, and constituent phases---to effective macroscopic properties. While these approaches have achieved remarkable success, they often require computationally intensive simulations and large design spaces that are difficult to explore exhaustively. Machine learning provides new opportunities to accelerate these workflows through surrogate modeling, dimensionality reduction, uncertainty-aware prediction, and efficient exploration of high-dimensional parameter spaces \cite{Bessa2017p633,Bessa2019p1904845}. Because materials design often relies on sparse experimental data, manufacturing variability, and multiscale structure--property relationships, probabilistic ML frameworks are increasingly used to quantify predictive confidence, guide data acquisition, and accelerate the discovery of materials with targeted properties \cite{Bessa2017p633,Allen2022,MLUQ2025p118061}.

A particularly active area concerns the development of data-driven relationships linking microstructure to mechanical performance. Such approaches support the discovery of novel materials, optimization of composite architectures, and the design of metamaterials or multifunctional systems with tailored properties. At the same time, recent advances in topology optimization have demonstrated how ML can accelerate structural design and enable rapid exploration of alternative configurations while preserving engineering constraints \cite{Chi2021p112739,Sanu2025,Senhora2022p115116,Vijayakumaran2025p106015}. These developments illustrate a broader trend in computational mechanics: AI is increasingly used to accelerate numerical simulation, construct surrogate models, assist design-space exploration, and support decision-making under uncertainty.

Despite these advances, the integration of ML with materials science and computational mechanics remains fundamentally dependent on physical knowledge. Constitutive behavior, conservation laws, manufacturing constraints, and multiscale interactions continue to provide essential structure for model development and validation. Consequently, many successful approaches combine ML with simulation, optimization, and physical modeling rather than treating these methodologies as competing alternatives \cite{Haywood-Alexander_Liu_Bacsa_Lai_Chatzi_2024}. This perspective aligns closely with the broader arguments developed in Sections~\ref{sec5} and~\ref{sec6}, where physical constraints and engineering knowledge are viewed as indispensable components of trustworthy AI systems.

\subsection{Scientific discovery and autonomous engineering workflows}

Recent developments suggest that AI may increasingly support not only isolated engineering tasks but also coordinated workflows involving simulation, optimization, data analysis, and design iteration. Agentic systems combine large language models with planning mechanisms, external tools, databases, simulation environments, and iterative feedback loops, enabling the execution of multi-step tasks rather than generating a single response. Such architectures are beginning to appear in scientific discovery platforms, autonomous laboratory concepts, and computer-aided engineering environments, where AI agents interact with numerical solvers, engineering software, and knowledge bases \cite{Jumper2021,Wang2023,Guo2026,Xi2025}.

In this context, scientific discovery refers not only to the automation of computational tasks but also to the generation, evaluation, and refinement of hypotheses through coordinated interaction between models, data, simulations, and experiments.

The growing interest in agentic AI stems from its potential to automate portions of scientific and engineering workflows that traditionally require extensive human coordination. Recent studies have explored collaborative multi-agent systems for scientific machine learning \cite{Jiang2026}, autonomous exploration of parameterized PDE spaces \cite{Vishwasrao2026}, and LLM-enabled computer-aided engineering frameworks that assist with model construction, simulation management, and design exploration \cite{Guo2026}. These developments suggest a gradual transition from isolated predictive models toward integrated computational ecosystems that combine simulation, optimization, data assimilation, UQ, and knowledge retrieval. In the long term, such capabilities may support autonomous laboratories, adaptive digital twins, self-driving simulation campaigns, and closed-loop engineering design environments. The U.S. Department of Energy Genesis Mission reflects the growing institutional interest in AI-enabled approaches for accelerating scientific and engineering productivity through tighter integration between computation, experimentation, and physical modeling \cite{DOEGenesis2025}.

At the same time, agentic systems do not eliminate the need for human expertise. Their effectiveness remains fundamentally dependent on scientific principles, engineering methodologies, validation procedures, and reasoning structures developed by human experts. Recent computer-aided engineering studies show that successful agentic workflows rely on mathematically rigorous reasoning templates and domain-specific knowledge encoded by engineers \cite{Guo2026}. Consequently, AI agents should be viewed as tools for augmenting engineering capability rather than replacing engineering judgment. Their successful deployment depends on the integration of physical knowledge, UQ, experimental validation, and professional oversight, reinforcing the central themes developed throughout this article.


\section{Knowledge formation, learning, and some limits of AI}
\label{sec8}

A fundamental question underlying the growing use of AI concerns the origin and long-term sustainability of the knowledge that feeds both natural and artificial intelligence: who will provide the knowledge for the “intelligence” in the medium and long term? This question is grounded in the understanding that AI is built upon the knowledge encoded and accumulated in large datasets. In this operational view, “intelligence” can be thought of as the ability to process information and apply it to solve problems. Intelligence, however, is not an automatic endowment; although it is a potential shared by all, it must be sought, developed, and cultivated over time.

From this perspective, intelligent behavior in natural systems emerges through several interconnected processes. Observation plays an important role by enabling the acquisition of knowledge—understood here as organized information—and the development of wisdom, understood as the capacity to apply knowledge effectively. Data collection, supported by study, research, and experience, contributes to the formation of behavioral models. Motivation can further shape learning trajectories, influenced by economic, ideological, cultural, or other interests. Discernment and judgment are also relevant components of adaptive decision-making, influenced by environmental factors such as social context and surrounding conditions, by interactions between rational and emotional processes, and by the capacity to improve future decisions through learning from past errors.

Computational systems commonly referred to as “artificial intelligence” employ algorithmic methods to analyze data and construct predictive or decision models that may approximate certain outcomes of these processes, without implying the presence of knowledge, understanding, or wisdom in the human sense.

Recent agentic systems do not alter this fundamental dependency. Even when capable of planning and executing multi-step workflows, such systems rely on scientific knowledge, engineering methodologies, and reasoning structures originally developed by human experts \cite{Guo2026,DOEGenesis2025}. Their performance remains linked to the continued production, validation, and transmission of engineering knowledge across generations.

There is no doubt about the importance and potential of AI systems in processing, organizing, and handling large amounts of information. In this sense, such systems can be regarded as powerful tools for supporting the search for knowledge, much as hard-copy encyclopedias served previous generations. However, both historical and contemporary knowledge systems rely fundamentally on the systematization of expertise produced by others. Neither humans nor machines are born knowledgeable. In humans, learning is a cognitive process that evolves throughout life through formal education, observation, experimentation, and evaluation. By contrast, ML refers to computational methods that analyze data and extract statistical patterns to produce models or make predictions, without implying cognition or understanding in the human sense.

A central concern is that future generations may become overly dependent on AI-facilitated knowledge searches, potentially overusing such tools. This dependence may reduce interest in formal education and limit opportunities for observation, experimentation, and critical evaluation. Education requires dedication, time, and effective teachers or mentors, while observation and experimentation depend on the execution of systematic tasks. Whether AI itself can truly possess discernment and judgment remains an open and debatable question.

This line of reasoning returns to the central issue raised at the outset: how will knowledge emerge and develop in the future, and who will produce the (data encoded) knowledge that informs intelligent systems in the medium and long term? Addressing this question is essential to ensure that the expanding use of AI strengthens rather than undermines the epistemological foundations of engineering education and practice. AI systems process and analyze large volumes of data, but the generation, interpretation, and validation of knowledge remain fundamentally human activities developed through education, observation, experimentation, and critical evaluation. Ensuring that these processes continue to be cultivated is therefore essential for the future creation and renewal of knowledge.

\section{Education and curriculum modernization}
\label{sec9}

The growing presence of AI in engineering practice has direct implications for the education and training of future mechanical engineers. As AI tools are integrated into design environments, simulation platforms, diagnostic systems, and technical communication workflows, engineers must develop the competencies to use these technologies responsibly and effectively \cite{Shneiderman2022HCAI}. 

This educational challenge is part of a broader global shift: two-thirds of countries now offer or plan to offer K--12 computer science education, often with increasing attention to AI-related content, indicating that AI literacy is progressively being recognized as a foundational competence rather than a specialized add-on \cite{stanford_ai_index_2025}. 

Evidence from higher education surveys also suggests that generative AI tools are already deeply embedded in students’ academic workflows. A recent survey of undergraduate students in the United Kingdom found that the vast majority reported using generative AI tools in their studies, frequently for tasks such as explaining concepts, summarizing academic material, and structuring ideas \cite{HEPI_AI_Survey_2026}. These trends reinforce the need for universities to provide structured guidance on responsible and effective AI use.

\begin{figure}[ht!]
\centering
\includegraphics[width=0.95\linewidth]{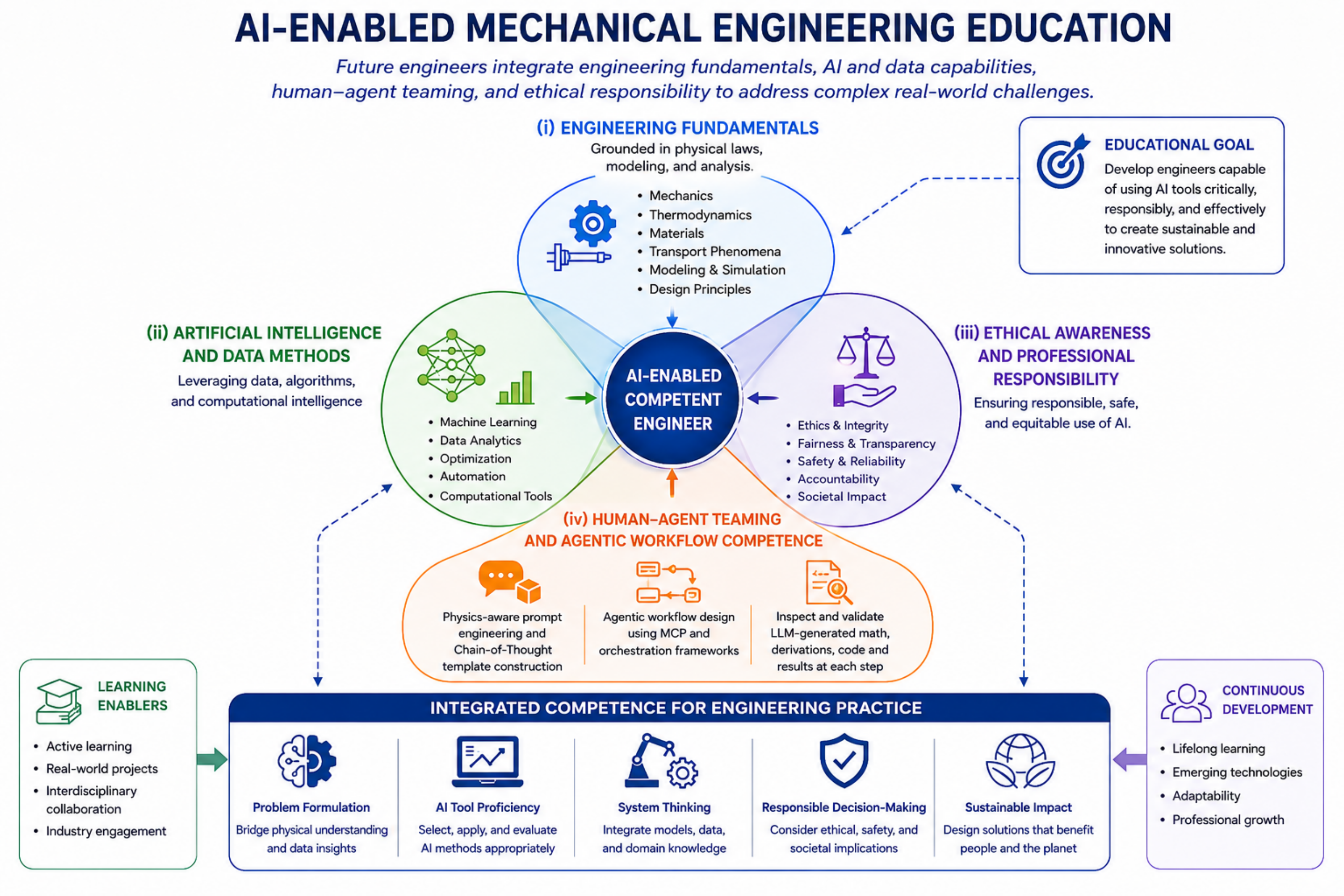}
\vspace{2mm}
\caption{Conceptual framework for AI-enabled ME education. Future engineers must integrate four complementary domains of competence: (i) strong engineering fundamentals grounded in physical laws, modeling, and analysis; (ii) artificial intelligence and data-driven methods; (iii) ethical awareness and professional responsibility; and (iv) human-agent teaming and agentic workflow supervision. The latter includes competencies such as physics-aware interaction with AI systems, supervision of AI-assisted workflows, inspection and validation of AI-generated analyses, and critical assessment of intermediate and final computational results. The integration of these domains enables the responsible, effective, and trustworthy use of AI in engineering practice.}
\label{fig:AI_education}
\end{figure}

Figure~\ref{fig:AI_education} presents a conceptual framework for AI-enabled ME education. The framework emphasizes the integration of four complementary domains of competence: engineering fundamentals, AI and data-driven methods, ethical awareness and professional responsibility, and human-agent teaming. Together, these domains provide the knowledge and skills required for engineers to use AI critically, responsibly, and effectively while maintaining scientific rigor, physical consistency, and professional accountability.

Industry evidence suggests that AI adoption is already reshaping workforce skill requirements \cite{McKinsey_State_of_AI_2025}. In addition to traditional engineering competencies, professionals increasingly require familiarity with data analysis, ML concepts, and digital engineering environments. However, technical knowledge alone is insufficient. Engineers must also cultivate capabilities that remain fundamentally human, including critical thinking, creativity, collaboration, ethical judgment, and the ability to interpret results within a broader physical and societal context.

Universities, therefore, face the challenge of modernizing engineering curricula to reflect this evolving landscape. Educational programs should incorporate interdisciplinary training that connects artificial intelligence, mechanical engineering, and systems thinking. Core topics may include ML fundamentals, hybrid modeling, physics-informed learning, symbolic reasoning, and data-driven engineering methods under physical constraints. Simultaneously, students must develop a clear understanding of the limitations of AI systems, including uncertainty, bias, interpretability, and appropriate domains of application.

This educational transformation is also reflected in national policy initiatives. The PBIA 2024--2028 \cite{PBIA2024} explicitly identifies workforce formation, digital literacy, and advanced AI training as strategic priorities, reinforcing the need for universities to prepare engineers capable of working at the intersection of physical systems, computational methods, and artificial intelligence.

Hands-on learning experiences are particularly important. Projects involving real-world datasets, experimental measurements, and simulation environments can help students bridge the gap between theoretical AI techniques and practical engineering problems. Such experiences also reinforce the importance of domain knowledge in interpreting AI-generated results.

Education must also address the ethical and professional dimensions of AI use. Engineers should be trained to recognize automation bias, critically evaluate algorithmic outputs, and maintain accountability when AI-assisted tools are used in engineering analyses or decisions. In this context, AI literacy should not be understood merely as technical proficiency, but as the ability to use intelligent tools while preserving scientific rigor and professional responsibility.

The emergence of agentic AI systems introduces an additional set of competencies related to human-AI collaboration. Future engineers may increasingly interact with AI agents capable of coordinating simulations, optimization routines, databases, and engineering software within multi-step workflows. In such settings, engineers must be able to formulate objectives, supervise workflow execution, inspect AI-generated code and mathematical reasoning, and verify the consistency of AI-generated results with physical laws, governing equations, and engineering requirements. These skills complement traditional engineering expertise and help ensure that increasing levels of automation remain consistent with physical principles, engineering requirements, and professional accountability.

At the same time, the growing presence of AI does not diminish the importance of experimental practice. Laboratory activities, prototyping, instrumentation, testing, and direct interaction with physical systems remain indispensable components of engineering education. These experiences cultivate physical intuition, practical judgment, and an understanding of uncertainty that cannot be acquired exclusively through digital tools or AI-mediated interactions.

Finally, continuing education will play an increasingly important role. As AI technologies evolve rapidly, lifelong learning opportunities — including professional training programs, short courses, and micro-credentials — will be essential to help practicing engineers stay current and integrate new digital tools into their work.


\section{Ethics, responsibility, and professional accountability}
\label{sec10}

The rapid diffusion of AI into engineering practice raises ethical questions that extend beyond abstract concerns about automation or employment \cite{Floridi2023EthicsAI}. In ME, AI systems increasingly contribute to reports, models, simulations, design decisions, and operational recommendations that can affect safety, reliability, economic outcomes, and public trust. This makes ethical responsibility inseparable from technical responsibility.

A central ethical challenge lies in the epistemic nature of contemporary AI systems. Generative and predictive models operate by identifying statistical regularities in data and producing outputs that are plausible according to learned patterns. They do not possess understanding, intent, or judgment, nor do they distinguish between correct and incorrect information in a human sense. As a result, AI systems may generate outputs that appear coherent, authoritative, and professionally formatted, yet be partially or entirely incorrect. This phenomenon, often described as “hallucination,” is not an anomaly but a structural consequence of statistical optimization. In engineering contexts, where plausibility cannot substitute for correctness, this creates a nontrivial ethical risk \cite{Rudin2019Interpretable}.

The risks associated with advanced AI systems arise from multiple sources, including malicious use, system malfunctions, and broader systemic impacts on institutions and societies \cite{InternationalAISafetyReport2026}. These categories highlight that responsible AI deployment requires both technical safeguards and institutional governance mechanisms capable of managing risks throughout the system life cycle.

The use of AI, therefore, introduces a clear responsibility gap if not properly governed. While AI tools can assist engineers, they cannot be moral or legal agents. Accountability for decisions, designs, and analyses remains unequivocally human. Delegating judgment to AI systems—explicitly or implicitly—undermines core principles of engineering ethics, including responsibility, due diligence, and commitment to public safety. This is particularly critical when AI-generated content is incorporated into technical documentation, feasibility studies, risk assessments, or safety-related decisions. This responsibility becomes even more important when AI systems participate in multi-step engineering workflows. In such settings, professional accountability extends beyond reviewing final outputs to include oversight of objectives, assumptions, intermediate decisions, validation procedures, and the conditions under which tasks are executed.

Another ethical dimension concerns the integrity of information and its misuse. The same technologies that enable productivity gains can also be used to generate misleading content at scale, including technically plausible but incorrect analyses, manipulated visualizations, or biased summaries. In engineering organizations, such misuse may not stem from malicious intent but from over-reliance on AI outputs under time or cost pressure. Ethical use of AI, therefore, requires not only safeguards against intentional abuse but also organizational cultures that encourage skepticism, verification, and traceability.

At the institutional level, the responsible use of AI requires governance mechanisms that align with engineering values. These include clear policies on acceptable use, documentation of AI-assisted workflows, validation procedures, and explicit human oversight in high-consequence applications \cite{FloridiCowls2019FivePrinciples,Shneiderman2020HCAI,Shneiderman2022HCAI}.

Ethical responsibility also intersects with broader societal concerns, such as data privacy, environmental impact, and equitable access to technology, reinforcing the need for transparency and accountability throughout the AI life cycle. 

This need is consistent with broader global trends: recent international assessments report growing policy attention to responsible AI, but also emphasize that practical mitigation efforts still lag behind the widespread recognition of risks such as inaccuracy, bias, and governance gaps \cite{stanford_ai_index_2025}. These concerns are further amplified when AI systems are deployed within organizational and market structures characterized by concentration of power, limited transparency, and weak mechanisms for contestation or remedy. In such contexts, questions of accountability involve not only technical reliability, but also who controls AI systems, who bears their risks, and who has the practical ability to challenge harmful outcomes \cite{AINow_ArtificialPower_2025}.

In summary, ethics in AI for ME is not an external constraint imposed on technology, but a core component of professional practice. Responsible AI use requires recognizing what AI systems are—and what they are not—while reaffirming the engineer’s role as the ultimate bearer of judgment, responsibility, and accountability. Only under these conditions can AI strengthen, rather than erode, the epistemic and ethical foundations of engineering.

Given the pace and impact of AI adoption, we call for coordinated action by the Brazilian professional regulation system—particularly the CREA and CONFEA councils—the Ministry of Education, research funding agencies, and academic institutions responsible for higher education. Such efforts should be aligned with broader national initiatives, including the PBIA 2024--2028 \cite{PBIA2024}, and encompass professional guidelines, codes of conduct, workforce development, and curricular modernization that address AI foundations, limitations, ethical considerations, and responsible use \cite{Floridi2023EthicsAI}.

\section{Validation, uncertainty quantification, and certification}
\label{sec11}

The increasing use of AI in ME raises a fundamental question: under what conditions can AI-generated predictions be trusted for engineering decision-making? Performance metrics obtained on benchmark datasets are often insufficient to answer this question. Engineering systems operate under uncertainty, evolving operating conditions, measurement noise, incomplete information, and potentially severe consequences of failure. As a result, validation, UQ, and certification become essential components of responsible AI deployment.

Validation seeks to establish whether a model is sufficiently accurate for its intended purpose. In engineering applications, this requires more than demonstrating predictive performance on historical data. Models must be evaluated under realistic operating conditions, tested for robustness to perturbations, and assessed for their ability to generalize beyond the specific datasets used during training. In many applications, validation must also consider consistency with known physical principles, conservation laws, and established engineering knowledge \cite{OberkampfRoy2010,Willcox2018}.

Uncertainty quantification plays a complementary role by characterizing the confidence associated with model predictions. Sources of uncertainty may arise from measurement errors, incomplete datasets, unknown parameters, model-form approximations, numerical discretization, or changing operating conditions. These sources are often classified as aleatory uncertainty, associated with inherent variability in physical systems, and epistemic uncertainty, associated with incomplete knowledge, limited data, or model-form inadequacy \cite{cunhajr_ekwaro2016,MLUQ2025p118061,Yi2025,GhanemHigdonOwhadi2017,Willcox2018,soize2017}.

AI systems often provide point predictions without explicitly communicating predictive confidence. For engineering applications, however, the uncertainty associated with a prediction may be as important as the prediction itself. Recent developments increasingly combine ML with probabilistic modeling, Bayesian inference, multi-fidelity methodologies, and surrogate modeling techniques to provide uncertainty estimates and support risk-informed decision-making \cite{GhanemHigdonOwhadi2017,soize2017}. Conformal prediction provide an alternative framework for constructing statistically valid prediction intervals with minimal distributional assumptions, offering a promising approach for uncertainty-aware AI systems in engineering applications \cite{Angelopoulos2023,Shafer2008,Vovk2022}.

The importance of UQ becomes particularly evident in digital twins, structural health monitoring, prognostics, predictive maintenance, and simulation-driven design. In these settings, decisions must often be made using imperfect information while balancing safety, performance, cost, and reliability. Quantified uncertainty enables engineers to assess predictive credibility, identify situations in which additional data are required, and avoid excessive confidence in AI-generated outputs \cite{Laflamme2026SHM,Willcox2018}.

Beyond validation and uncertainty estimation lies the challenge of certification. Traditional engineering systems are frequently subject to qualification, verification, and certification procedures established by industry standards and regulatory bodies. AI-based systems introduce additional difficulties because their behavior may depend on training data, model updates, and operational environments. Consequently, certification cannot rely solely on traditional software verification approaches. Emerging frameworks increasingly emphasize continuous monitoring, traceability, documentation of model development, uncertainty reporting, and explicit human oversight throughout the model life cycle \cite{NIST_AIRMF_2023, ISO_IEC_23894_2023}.

Although regulatory frameworks vary across sectors such as aerospace, automotive, nuclear energy, and pressure-vessel design, a common challenge emerges when AI is incorporated into engineering workflows: establishing sufficient evidence that system performance remains reliable, traceable, auditable, and robust throughout its operational life cycle. Bridging the gap between AI capabilities and engineering certifiability is likely to become a central challenge for deploying AI in safety-critical ME applications.

It is also important to distinguish between exploratory and operational uses of AI. Systems intended for scientific exploration, hypothesis generation, preliminary design studies, or research support may tolerate higher levels of uncertainty and lower levels of formal verification. By contrast, systems intended for operational deployment in safety-critical environments require substantially stronger evidence regarding robustness, reliability, traceability, and performance under uncertainty. The degree of validation should therefore be commensurate with the consequences of using the model.

Ultimately, trustworthy AI in ME requires a hierarchy of evidence. Predictive accuracy alone is insufficient. Reliable deployment depends on rigorous validation, quantified uncertainty, physical consistency, transparent documentation, and appropriate governance mechanisms. Together, these elements provide the foundation for integrating AI systems into engineering practice while preserving safety, reliability, and professional accountability.


\section{Conclusions}
\label{sec12}

Artificial intelligence is no longer a peripheral tool in mechanical engineering; it is progressively becoming embedded in modeling environments, design workflows, operational systems, and professional communication. Yet, as this manuscript has argued, its integration into engineering practice cannot be evaluated solely on the basis of performance gains or automation potential. Mechanical engineering operates in the physical world, under constraints imposed by energy, materials, uncertainty, safety, and accountability. The decisive question, therefore, is not whether AI will transform engineering, but whether it will do so in a manner that reinforces—rather than weakens—the epistemic, technical, and ethical foundations of the profession.

A central insight emerging from the round-table contributions is that AI must be understood as a component within engineered systems, not as an autonomous substitute for engineering reasoning. Contemporary AI systems excel at statistical approximation and pattern extraction, but they remain bounded by data distributions, computational resources, and training objectives. When deployed in high-consequence contexts—such as structural integrity assessment, dynamic system modeling, energy infrastructure design, or digital twins—their outputs must be constrained by physical laws, validated against domain knowledge, and embedded within transparent decision-making processes. The integration of physics-informed and hybrid learning frameworks is therefore not a refinement, but a structural requirement for reliability, interpretability, and extrapolation beyond training regimes.

AI’s expansion has material consequences for energy demand and infrastructure pressures. Large-scale model training and inference rely on data centers with high power density and demanding cooling requirements. Responsible AI in engineering must therefore include energy awareness as a first-class design variable. Mechanical engineers—historically central to energy conversion, thermal management, and efficiency—have both the competence and the responsibility to address this challenge. Portfolio-based energy solutions, efficiency improvements, and system-level optimization are likely to define the sustainability trajectory of AI infrastructure. In this sense, the environmental footprint of computation is not external to engineering judgment; it is part of it.

Beyond technical and infrastructural issues lies a deeper epistemic concern: AI systems are trained on accumulated knowledge, yet the long-term vitality of engineering depends on continued human observation, experimentation, mentorship, and critical reasoning. Overreliance on automated outputs risks narrowing intellectual engagement and weakening the generative processes that produce new knowledge. AI can amplify access to information, but it cannot replace discernment, contextual understanding, or ethical judgment. Preserving and strengthening the human foundations of engineering knowledge is therefore inseparable from the responsible use of AI tools.

Ethical and professional accountability remains unequivocally human. No matter how capable they become, AI systems are not bearers of professional duty, legal accountability, or ethical judgment. In ME, where decisions can affect safety, economic stability, and public welfare, delegation of judgment to opaque models without verification constitutes a breach of professional duty. Transparency of use, traceability of AI-assisted workflows, explicit human oversight, and institutional governance are essential conditions for the legitimate deployment of AI. The modernization of engineering curricula and the establishment of clear professional guidelines—including codes of conduct for AI use—are urgent measures, not optional enhancements.

Taken together, the discussions presented in this article suggest a set of integrating principles for the constructive adoption of AI in ME:
\begin{enumerate}
    \item \textbf{Human–Agent Teaming:} Future engineering practice will increasingly involve collaboration between human experts and AI agents capable of coordinating simulations, optimization routines, databases, and engineering software. Developing competencies for supervising, validating, and directing such workflows will become an essential component of engineering practice.
    \item \textbf{Physics as constraint and guide:} Learning systems deployed in engineering must be anchored in physical laws, structural knowledge, and domain constraints.
    \item \textbf{Energy-aware computation:} The design and deployment of AI systems must account for their energy, carbon, and infrastructure implications.
    \item \textbf{Validation and uncertainty awareness:} AI systems used in engineering must be accompanied by appropriate validation, uncertainty quantification, and life-cycle monitoring procedures commensurate with the consequences of their use.
    \item \textbf{Human accountability:} Responsibility for decisions, documentation, and safety remains with licensed professionals and institutions.
    \item \textbf{Educational renewal:} Engineering education must integrate AI literacy, critical reasoning, and ethical reflection as core competencies.
\end{enumerate}

In the immediate term, these principles call for careful governance of AI-assisted workflows and explicit integration of verification practices. In the medium term, they motivate research into hybrid, physics-informed, and computationally efficient methodologies that support reliable digital twins and dynamic-system diagnostics. In the long term, they point to the need for cultural and institutional adaptation—ensuring that knowledge formation, engineering training, and infrastructure stewardship evolve in parallel with AI capabilities. If integrated with rigor and foresight, AI may help expand ME's analytical reach, shorten development cycles, improve decision support, and enable new forms of scientific and technological innovation.

Artificial intelligence will undoubtedly remain a transformative force. But the future of AI in ME will ultimately depend not on how much can be automated, but on how effectively human expertise, physical understanding, and computational intelligence are integrated into a coherent engineering practice. When embedded within a framework of physical consistency, energy awareness, institutional responsibility, and human judgment, AI can serve as a powerful enabler of engineering excellence. Absent such a framework, it risks amplifying fragility and obscuring accountability. The defining challenge of the coming decade is therefore not technological acceleration, but principled integration.


\section*{Acknowledgments}

The authors gratefully acknowledge the Brazilian Society of Mechanical Sciences and Engineering (ABCM) for organizing COBEM 2025 and for fostering the Round Table on Artificial Intelligence that motivated this article. The authors also thank Prof. Ibrahim Dincer (University of Ontario Institute of Technology) and Mr. Antonio Nabholz (Versatus HPC) for their valuable participation in the Round Table and for the insightful perspectives shared during the discussions. The constructive comments and criticisms from the anonymous referees are also appreciated.


\section*{Funding}

This work was supported by the Brazilian National Council for Scientific and Technological Development (CNPq) under grants 305476/2022-0 (A. Cunha Jr) and 306053/2022-5 (A. Scotti); by the Carlos Chagas Filho Research Foundation of the State of Rio de Janeiro (FAPERJ) under grants 211.037/2019 and 204.477/2024 (A. Cunha Jr); and by the Coordination for the Improvement of Higher Education Personnel (CAPES), Finance Code 001. J. V. C. Vargas acknowledges support from CNPq under projects 443823/2018-9, 446787/2020-5, 313646/2020-1, 308460/2020-0, 312516/2019-3, 408073/2021-7, 405864/2022-1, and 408080/2022-1; from CAPES (Ministry of Education, Brazil) under project CAPES-PRINT-UFPR-88881.311981/2018-01; from the Araucaria Foundation of Paraná, Brazil, under project 115/2018 (no. 50.579 – PRONEX); and from the Research \& Development Foundation (FUNDEP – ROUTE 2030) under project 27192.22. C. Nataraj gratefully acknowledges support from the U.S. Office of Naval Research under Grant No. N00014-22-1-2480 (Program Manager: Capt. L. J. Peterson).

\section*{Declarations}

\noindent
\textbf{Conflict of Interest:} The authors declare that they have no conflict of interest.\\

\noindent
\textbf{Data availability:} No datasets were generated or analyzed during the current study.\\

\noindent
\textbf{Code availability:} Not applicable.\\

\noindent
\textbf{Disclaimer:} Artificial intelligence tools, including ChatGPT and Grammarly, were employed to assist with language polishing, stylistic improvements, and iterative image generation. The authors assume full responsibility for the intellectual content, structure, and final version of the manuscript.

\bibliographystyle{spbasic} 

\bibliography{References} 

\end{document}